\documentclass[
]{ceurart}

\usepackage{listings}
\usepackage{tikz}
\usepackage{amsfonts}
\usepackage{amsthm}
\usetikzlibrary{arrows}
\usetikzlibrary{calc}
\usepackage{algorithm2e}
\RestyleAlgo{ruled}
\usepackage{comment}

\usepackage{graphicx}
\usepackage{subcaption}
\usepackage{mwe}

\begin{document}

\copyrightyear{2022}
\copyrightclause{Copyright for this paper by its authors.
  Use permitted under Creative Commons License Attribution 4.0
  International (CC BY 4.0).}

\conference{Woodstock'22: Symposium on the irreproducible science,
  June 07--11, 2022, Woodstock, NY}

\title{Learning Rules from KGs Guided by Language Models}





\author[1,2]{Zihang Peng}[
email=zp321@ic.ac.uk,
]
\cormark[1]
\author[2]{Daria Stepanova}[
email=Daria.Stepanova@de.bosch.com,
]
\cormark[1]
\author[3]{Vinh Thinh Ho}[
email=hovnh@amazon.com,
]
\author[2]{Heike Adel}[
email=Heike.Adel@de.bosch.com,
]
\author[1]{Alessandra Russo}[
email=a.russo@ic.ac.uk,
]
\author[2]{Simon Ott}[
email=Simon.Ott@de.bosch.com,
]
\address[1]{Imperial College London, London, UK}
\address[2]{Bosch Center for Artificial Intelligence, Renningen, Germany}
\address[3]{Amazon Research, Berlin, Germany}

\cortext[cor1]{Corresponding author. pzh1997@outlook.com}

\begin{abstract}
Advances in information extraction have enabled the automatic construction of large knowledge graphs (e.g., Yago, Wikidata or Google KG), which are widely used in many applications like semantic search or data analytics. However, due to their semi-automatic construction, KGs are often incomplete.
Rule learning methods, concerned with the extraction of frequent patterns from KGs and casting them into rules, can be applied to predict potentially missing facts. A crucial step in this process is
rule ranking. Ranking of rules is especially challenging over highly incomplete or biased KGs (e.g., KGs predominantly storing facts about famous people), as in this case biased rules might fit the data best and be ranked at the top based on standard statistical metrics like rule confidence. To address this issue, prior works proposed to rank rules not only relying on the original KG but also facts predicted by a KG embedding model. At the same time, with the recent rise of Language Models (LMs), several works have claimed that LMs can be used as alternative means for KG completion. In this work, our goal is to verify to which extent the exploitation of LMs is helpful for improving the quality of rule learning systems. 
\end{abstract}

\maketitle

\section{Introduction}\label{sec:intro}
Knowledge Graphs (KGs) store large collections of factual knowledge, and they are widely used in various applications including semantic search or data analytics. 

However, due to their semi-automatic construction, KGs might be incomplete. Rules of the form $\mathit{head}\leftarrow \mathit{body}$, where $\mathit{head}$ is an atom and  $\mathit{body}$ is a conjunction of atoms, are learnt from KGs, ranked based on statistical metrics like confidence~\cite{goethals2002relational} and applied to deduce potentially missing facts. As traditional statistical metrics (e.g., confidence) have been originally designed for evaluating rules extracted from complete data, relying on them when ranking rules learnt from incomplete KGs might be misleading, resulting in incorrect predictions~\cite{ho2018rule}. For instance, the rule $\mathit{r1:livesIn(X,Z)}\leftarrow \mathit{politicialOf(X,Y), capitalOf(Y,Z)}$ stating that politicians typically live in capitals  extracted from an incomplete KG storing facts about only very famous politicians will have high confidence and support, introducing a clear bias. 

To address the issue of biased rules and inaccurate predictions associated with them, several works have proposed to combine rule learning and KG embeddings in various ways~\cite{
DBLP:conf/aib/BoschinJKS22,DBLP:conf/akbc/MeilickeBS21,DBLP:journals/tkde/OmranWW21,
DBLP:conf/www/ZhangPWCZZBC19,ho2018rule}. At the same time, with the recent rise of language models (LMs)~\cite{bengio2000neural,dauphin2017language}, there have been also many discussions regarding the possibilities of employing LMs for KG-related tasks~\cite{petroni2019language,DBLP:journals/corr/abs-2204-06031,DBLP:journals/corr/abs-2010-11967}. A very recent work~\cite{DBLP:conf/acl/JiangAJW0023} has shown that LMs can be in principle utilized for KG completion, leading to promising results that are often better than those of existing KG embedding models.

In this preliminary work, we evaluate to which extent language models used out of the box can be helpful for improving the rule learning process \footnote{https://github.com/pzh97/Learning-Rules-from-KGs-Guided-by-Language-Models}. More specifically, we develop a system prototype\footnote{https://tinyurl.com/yckuds8b} that exploits large pre-trained language models for ranking rules learnt from KGs in the spirit of~\cite{ho2018rule}. More specifically, similar to~\cite{ho2018rule}, given a knowledge graph, we extract Horn rules from it, and then rank them using the hybrid confidence metric, which accounts not only for the descriptive quality of the rules (like confidence~\cite{goethals2002relational} or its variations~\cite{DBLP:journals/vldb/GalarragaTHS15}), but also predictive quality. However, in contrast to~\cite{ho2018rule}, for evaluating the predictive quality of the rules, we make use of a pre-trained language model rather than a knowledge graph embedding model. Our preliminary experiments on the Wiki44K dataset \cite{ho2018rule} with a popular Bert LM without fine-tuning have shown a potential for using LMs for rule evaluation. 
\vspace{-.3cm}
\section{System Overview}
\begin{figure}[t]
\includegraphics[scale=0.39]{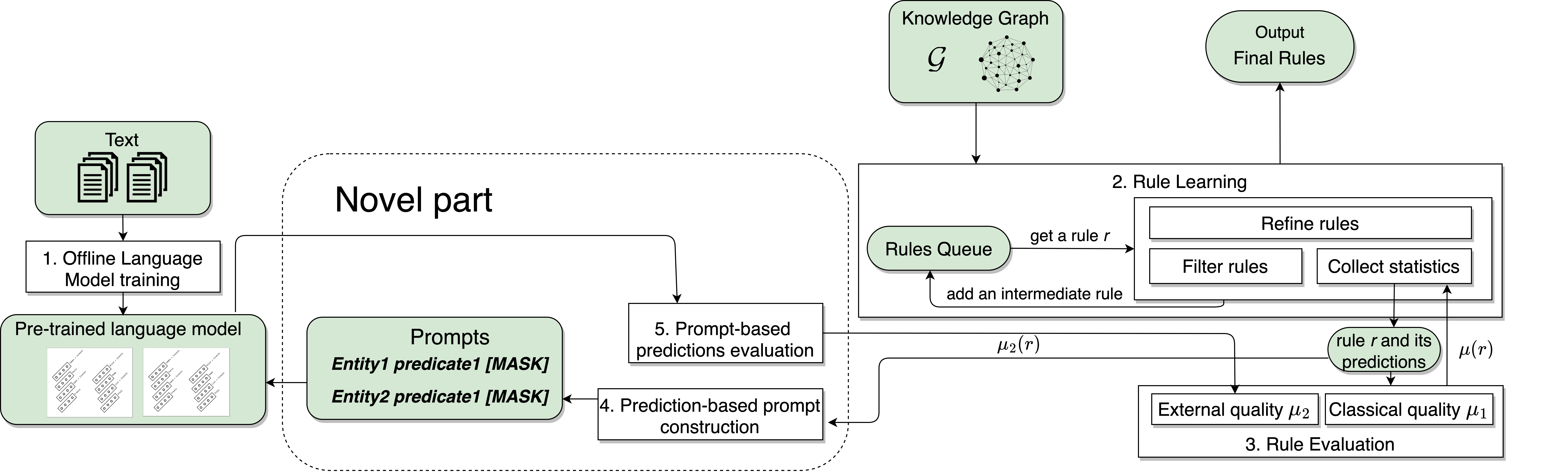}
\caption{System Overview.}
\label{system}
\end{figure}

We have implemented the method for learning rules with the help of language models in a system prototype. The overview of the system is given in Figure~\ref{system}. In the first step, a language model $\mathcal{L}$ is trained on a large textual corpus, or alternatively any pre-trained model can be utilized, which makes our system flexible. Second, rules are extracted from a knowledge graph using association rule mining similar to the standard rule learning methods like~\cite{DBLP:conf/ijcai/MeilickeCRS19,DBLP:journals/vldb/GalarragaTHS15}. Third, the rules are ranked using a hybrid rule quality function, which is a combination of an established statistical quality measurement $\mu_1$  (e.g., confidence~\cite{goethals2002relational}, i.e., conditional probability of the rule head given its body, or its variations~\cite{DBLP:journals/vldb/GalarragaTHS15}) and an external score $\mu_2$ generated relying on the pre-trained language model $\mathcal{L}$.  
 The rule evaluation formula~\cite{ho2018rule} combines these scores using a weighted average as follows:
\begin{align} 
    \mu(r) = (1-\lambda) * \mu_1(r) + \lambda * \mu_2(r)
    \label{eq:hyb}
\end{align}
where $\lambda \in [0, 1]$ is the weight of the language model-based score. 
The external rule quality measurement $\mu_2$ is the reciprocal rank 
of the correct answer assigned by the language model 
$\mathcal{L}$ given the head and relation of a triple predicted by a rule $r$ as a query.

More specificlly, for each mined rule considered individually, we collect all of its predictions and construct an LM prompt. For example, given the prediction $\mathit{livesIn(john,budapest)}$ produced by the rule $\mathit{r1}$ from Sec.~\ref{sec:intro}, we construct the prompt \texttt{``John lives in [MASK]''}. We then retrieve the predictions produced by the language model, and only in the case if \texttt{Budapest} appears in the top $n$ (user-specified parameter) results, we treat the fact $\mathit{livesIn(john,budapest)}$ as the correct one and do not penalize $\mathit{r1}$.
The score $\mu_2$ is then 
computed for each rule as the reciprocal rank of rule predictions that are confirmed to be correct by the language model (i.e., among top-$n$ results returned by LM for the respective masked query). 

Rules ranked relying on $\mu$ are then applied to obtain fact predictions in the standard way~\cite{DBLP:journals/vldb/GalarragaTHS15}.

\begin{figure*}[t]
     \centering
     \begin{subfigure}[b]{0.475\textwidth}
         \centering
         \includegraphics[width=\textwidth]{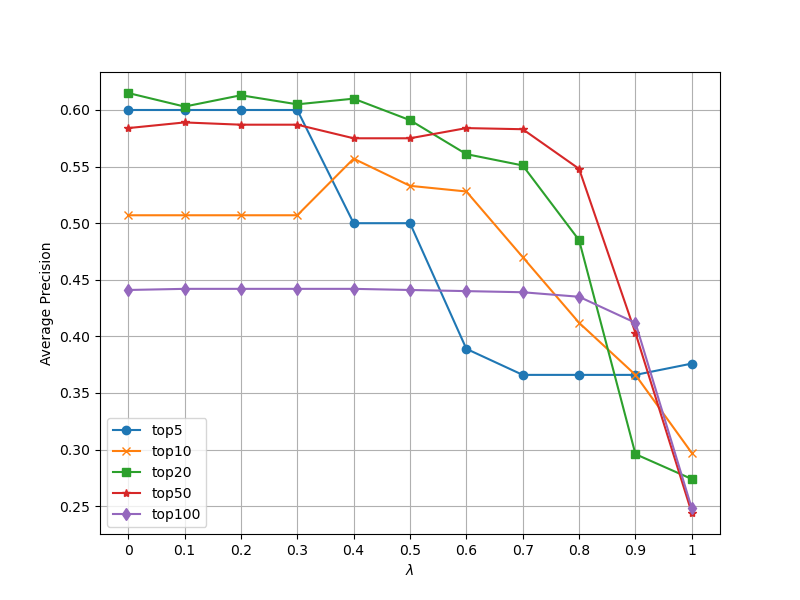}
         \caption{BERT, Standard Confidence}
         \label{fig: bert-standard confidence}
     \end{subfigure}
     \hfill
     \begin{subfigure}[b]{0.475\textwidth}
         \centering
         \includegraphics[width=\textwidth]{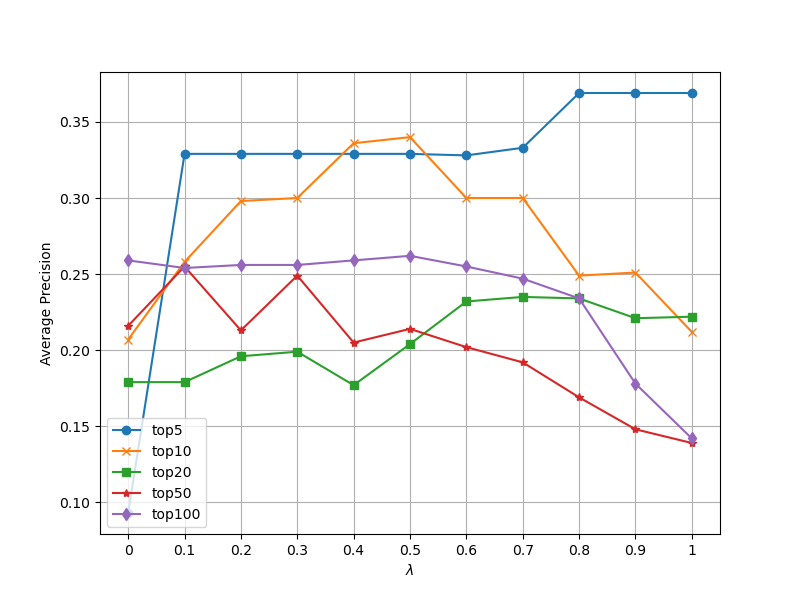}
         \caption{BERT, PCA}
         \label{fig: bert-pca}
     \end{subfigure}
        \caption{Average precision of predictions made by rules computed relying on the language model Bert.}
        \label{fig:bert}
\end{figure*}

\vspace{-.3cm}
\section{Evaluation}
\paragraph{Experimental Setup.} We have implemented our approach in a system prototype and make the code available\footnote{https://tinyurl.com/yckuds8b}.
Preliminary evaluation of our approach has been performed on the KG benchmark \emph{Wiki44k}, containing 250,000 binary facts, 100 relations, and 44,000 entities. The original KG is considered to be the test KG $\mathcal{G}_{\mathit{test}}$, and the train KG $\mathcal{G}$ is obtained from the test KG by discarding 20\% of facts for each relation. 
Following~\cite{ho2018rule}, the quality of  rules is estimated as the average percentage of facts predicted by a rule that are in the test KG. 
\begin{itemize}
    \item $pred\_prec_{CW}(r)=\frac{|\mathcal{G}_r\cap\mathcal{G}_{\mathit{test}}\setminus \mathcal{G}|}{|\mathcal{G}_r \setminus \mathcal{G}|}$, where $\mathcal{G}_r$ is the set of facts deduced by a given rule $r$. 
    \item $pred\_prec_{CW}(R)=\frac{\sum_{r\in R}pred\_prec_{CW}(r)}{|R|}$, where $R$ is a set of rules. 
\end{itemize}

We learned rules of form $h(X, Z) \leftarrow p(X, Y ), q(Y, Z)$ from $\mathcal{G}$ with a support of $\geq10$, standard confidence of $\geq0.1$ and a head coverage of $\geq0.01$. Then, we rank these rules using Equation~\ref{eq:hyb} with $\lambda \in \{0, 0.1, 0.2, \ldots, 1\}$. 
For the descriptive part of the score $\mu_1$ we used the standard metrics, i.e., confidence~\cite{goethals2002relational} and PCA confidence~\cite{DBLP:journals/vldb/GalarragaTHS15}.
For the external score $\mu_2$, we used  language model BERT and masked object entities in triples as queries. We focused on single token objects 
and mapped multi-token KG entities to single tokens in the tokenizer vocabulary of BERT using sentence embeddings and cosine similarity. Unused tokens (e.g., stop words) were omitted in the rankings for each query. 
We have chosen Bert without any fine-tuning on KGs as the LM, but our code allows one to conveniently integrate also other LMs.

\paragraph{Results.} In Figure~\ref{fig:bert} we report the average precision of the top-$k$ rules for various values of $\lambda$, with $k \in \{5,10,20,50,100\}$. The standard confidence (resp. PCA confidence) corresponds to $\lambda=0$. In case of $\lambda=1$, the ranking of the rules is done exclusively relying on the LM. 

We observe that in case of the standard confidence, the integration of a language model results in some positive effect for top 10 rules and $\lambda=0.4$ as well as for top 50 rules and $\lambda=0.6$.
In case of a PCA confidence, there is a boost in precision of the resulting predictions mainly for $\lambda$ between $0.4$ and $0.6$. For top-5 rules, even relying exclusively on LMs when ranking rules yields positive effect. 
\vspace{-.3cm}
\section{Conclusion}
We have proposed and implemented a method for making use of language models in the rule learning process, in particular at the rule ranking stage. The initial experiments show the potential of the method even when a language model is not fine-tuned on the original KG. We believe that when coupled with any other existing rule learning systems, e.g., AnyBurl~\cite{DBLP:conf/ijcai/MeilickeCRS19} or AMIE~\cite{DBLP:journals/vldb/GalarragaTHS15} which support  more expressive rule language bias, our method can demonstrate improvements in the quality of the learnt rules and accuracy of predictions they produce.

\vspace{-.3cm}
\bibliography{sample-ceur}

\begin{thebibliography}{14}
\expandafter\ifx\csname natexlab\endcsname\relax\def\natexlab#1{#1}\fi
\providecommand{\url}[1]{\texttt{#1}}
\providecommand{\href}[2]{#2}
\providecommand{\path}[1]{#1}
\providecommand{\DOIprefix}{doi:}
\providecommand{\ArXivprefix}{arXiv:}
\providecommand{\URLprefix}{URL: }
\providecommand{\Pubmedprefix}{pmid:}
\providecommand{\doi}[1]{\href{http://dx.doi.org/#1}{\path{#1}}}
\providecommand{\Pubmed}[1]{\href{pmid:#1}{\path{#1}}}
\providecommand{\bibinfo}[2]{#2}
\ifx\xfnm\relax \def\xfnm[#1]{\unskip,\space#1}\fi
\bibitem[{Goethals and Bussche(2002)}]{goethals2002relational}
\bibinfo{author}{B.~Goethals}, \bibinfo{author}{J.~V.~d. Bussche},
\newblock \bibinfo{title}{Relational association rules: Getting warmer},
\newblock in: \bibinfo{booktitle}{Pattern Detection and Discovery},
  \bibinfo{publisher}{Springer}, \bibinfo{year}{2002}, pp.
  \bibinfo{pages}{125--139}.
\bibitem[{Ho et~al.(2018)Ho, Stepanova, Gad-Elrab, Kharlamov, and
  Weikum}]{ho2018rule}
\bibinfo{author}{V.~T. Ho}, \bibinfo{author}{D.~Stepanova},
  \bibinfo{author}{M.~H. Gad-Elrab}, \bibinfo{author}{E.~Kharlamov},
  \bibinfo{author}{G.~Weikum},
\newblock \bibinfo{title}{Rule learning from knowledge graphs guided by
  embedding models},
\newblock in: \bibinfo{booktitle}{{ISWC} 2018}, \bibinfo{year}{2018}, pp.
  \bibinfo{pages}{72--90}.
\bibitem[{Boschin et~al.(2022)Boschin, Jain, Keretchashvili, and
  Suchanek}]{DBLP:conf/aib/BoschinJKS22}
\bibinfo{author}{A.~Boschin}, \bibinfo{author}{N.~Jain},
  \bibinfo{author}{G.~Keretchashvili}, \bibinfo{author}{F.~M. Suchanek},
\newblock \bibinfo{title}{Combining embeddings and rules for fact prediction},
\newblock in: \bibinfo{booktitle}{{AIB} 2022}, volume~\bibinfo{volume}{99} of
  \textit{\bibinfo{series}{OASIcs}}, \bibinfo{year}{2022}, pp.
  \bibinfo{pages}{4:1--4:30}.
\bibitem[{Meilicke et~al.(2021)Meilicke, Betz, and
  Stuckenschmidt}]{DBLP:conf/akbc/MeilickeBS21}
\bibinfo{author}{C.~Meilicke}, \bibinfo{author}{P.~Betz},
  \bibinfo{author}{H.~Stuckenschmidt},
\newblock \bibinfo{title}{Why a naive way to combine symbolic and latent
  knowledge base completion works surprisingly well},
\newblock in: \bibinfo{booktitle}{{AKBC} 2021}, \bibinfo{year}{2021}.
\bibitem[{Omran et~al.(2021)Omran, Wang, and
  Wang}]{DBLP:journals/tkde/OmranWW21}
\bibinfo{author}{P.~G. Omran}, \bibinfo{author}{K.~Wang},
  \bibinfo{author}{Z.~Wang},
\newblock \bibinfo{title}{An embedding-based approach to rule learning in
  knowledge graphs},
\newblock \bibinfo{journal}{{IEEE} Trans. Knowl. Data Eng.}
  \bibinfo{volume}{33} (\bibinfo{year}{2021}) \bibinfo{pages}{1348--1359}.
\bibitem[{Zhang et~al.(2019)Zhang, Paudel, Wang, and \textit{et
  al}}]{DBLP:conf/www/ZhangPWCZZBC19}
\bibinfo{author}{W.~Zhang}, \bibinfo{author}{B.~Paudel},
  \bibinfo{author}{L.~Wang}, \bibinfo{author}{J.~C. \textit{et al}},
\newblock \bibinfo{title}{Iteratively learning embeddings and rules for
  knowledge graph reasoning},
\newblock in: \bibinfo{booktitle}{{WWW}}, \bibinfo{year}{2019}, pp.
  \bibinfo{pages}{2366--2377}.
\bibitem[{Bengio et~al.(2000)Bengio, Ducharme, and Vincent}]{bengio2000neural}
\bibinfo{author}{Y.~Bengio}, \bibinfo{author}{R.~Ducharme},
  \bibinfo{author}{P.~Vincent},
\newblock \bibinfo{title}{A neural probabilistic language model},
\newblock \bibinfo{journal}{Advances in neural information processing systems}
  \bibinfo{volume}{13} (\bibinfo{year}{2000}).
\bibitem[{Dauphin et~al.(2017)Dauphin, Fan, Auli, and
  Grangier}]{dauphin2017language}
\bibinfo{author}{Y.~N. Dauphin}, \bibinfo{author}{A.~Fan},
  \bibinfo{author}{M.~Auli}, \bibinfo{author}{D.~Grangier},
\newblock \bibinfo{title}{Language modeling with gated convolutional networks},
\newblock in: \bibinfo{booktitle}{International conference on machine
  learning}, \bibinfo{organization}{PMLR}, \bibinfo{year}{2017}, pp.
  \bibinfo{pages}{933--941}.
\bibitem[{Petroni et~al.(2019)Petroni, Rockt{\"a}schel, Lewis, Bakhtin, Wu,
  Miller, and Riedel}]{petroni2019language}
\bibinfo{author}{F.~Petroni}, \bibinfo{author}{T.~Rockt{\"a}schel},
  \bibinfo{author}{P.~Lewis}, \bibinfo{author}{A.~Bakhtin},
  \bibinfo{author}{Y.~Wu}, \bibinfo{author}{A.~H. Miller},
  \bibinfo{author}{S.~Riedel},
\newblock \bibinfo{title}{Language models as knowledge bases?},
\newblock \bibinfo{journal}{arXiv preprint arXiv:1909.01066}
  (\bibinfo{year}{2019}).
\bibitem[{AlKhamissi et~al.(2022)AlKhamissi, Li, Celikyilmaz, Diab, and
  Ghazvininejad}]{DBLP:journals/corr/abs-2204-06031}
\bibinfo{author}{B.~AlKhamissi}, \bibinfo{author}{M.~Li},
  \bibinfo{author}{A.~Celikyilmaz}, \bibinfo{author}{M.~Diab},
  \bibinfo{author}{M.~Ghazvininejad},
\newblock \bibinfo{title}{A review on language models as knowledge bases},
\newblock \bibinfo{journal}{CoRR} \bibinfo{volume}{abs/2204.06031}
  (\bibinfo{year}{2022}).
\bibitem[{Wang et~al.(2020)Wang, Liu, and
  Song}]{DBLP:journals/corr/abs-2010-11967}
\bibinfo{author}{C.~Wang}, \bibinfo{author}{X.~Liu}, \bibinfo{author}{D.~Song},
\newblock \bibinfo{title}{Language models are open kgs},
\newblock \bibinfo{journal}{CoRR} \bibinfo{volume}{abs/2010.11967}
  (\bibinfo{year}{2020}).
\bibitem[{Jiang et~al.(2023)Jiang, Agarwal, Jin, Wang, Sun, and
  Han}]{DBLP:conf/acl/JiangAJW0023}
\bibinfo{author}{P.~Jiang}, \bibinfo{author}{S.~Agarwal},
  \bibinfo{author}{B.~Jin}, \bibinfo{author}{X.~Wang},
  \bibinfo{author}{J.~Sun}, \bibinfo{author}{J.~Han},
\newblock \bibinfo{title}{Text augmented open knowledge graph completion via
  pre-trained language models},
\newblock in: \bibinfo{booktitle}{{ACL} 2023}, \bibinfo{year}{2023}, pp.
  \bibinfo{pages}{11161--11180}.
\bibitem[{Gal{\'{a}}rraga et~al.(2015)Gal{\'{a}}rraga, Teflioudi, Hose, and
  Suchanek}]{DBLP:journals/vldb/GalarragaTHS15}
\bibinfo{author}{L.~Gal{\'{a}}rraga}, \bibinfo{author}{C.~Teflioudi},
  \bibinfo{author}{K.~Hose}, \bibinfo{author}{F.~M. Suchanek},
\newblock \bibinfo{title}{Fast rule mining in ontological knowledge bases with
  {AMIE+}},
\newblock \bibinfo{journal}{{VLDB} J.} \bibinfo{volume}{24}
  (\bibinfo{year}{2015}) \bibinfo{pages}{707--730}.
\bibitem[{Meilicke et~al.(2019)Meilicke, Chekol, Ruffinelli, and
  Stuckenschmidt}]{DBLP:conf/ijcai/MeilickeCRS19}
\bibinfo{author}{C.~Meilicke}, \bibinfo{author}{M.~W. Chekol},
  \bibinfo{author}{D.~Ruffinelli}, \bibinfo{author}{H.~Stuckenschmidt},
\newblock \bibinfo{title}{Anytime bottom-up rule learning for knowledge graph
  completion},
\newblock in: \bibinfo{booktitle}{{IJCAI}}, \bibinfo{publisher}{ijcai.org},
  \bibinfo{year}{2019}, pp. \bibinfo{pages}{3137--3143}.

\end{thebibliography}

\end{document}